\documentclass[runningheads]{llncs}
\usepackage{graphicx}

\usepackage{tikz}
\usepackage{comment}
\usepackage{amsmath,amssymb} 
\usepackage{color}
\usepackage{cite}
\usepackage[accsupp]{axessibility}  

\usepackage{multirow}
\usepackage{booktabs}
\usepackage{subfig}
\usepackage{bbm}

\begin{document}
\pagestyle{headings}
\mainmatter
\def\ECCVSubNumber{1762}  

\title{MoDA: Map style transfer for self-supervised Domain Adaptation of embodied agents} 

\titlerunning{MoDA}
%
\author{Eun Sun Lee\inst{1}\orcidID{0000-0003-1731-5714} \and
Junho Kim\inst{1}\orcidID{0000-0002-5947-2147} \and  
SangWon Park\inst{1}\orcidID{0000-0002-9735-1303} \and 
Young Min Kim\inst{1}\orcidID{0000-0002-6735-8539}}
\authorrunning{E. Lee et al.}
%
\institute{Seoul National University, Seoul, 08826, Republic of Korea \\
\email{ \{ eunsunlee, 82magnolia,  paulmoguri, youngmin.kim \} @snu.ac.kr}\\
}
\maketitle

\begin{abstract}
We propose a domain adaptation method, MoDA, which adapts a pretrained embodied agent to a new, noisy environment without ground-truth supervision.
Map-based memory provides important contextual information for visual navigation, and exhibits unique spatial structure mainly composed of flat walls and rectangular obstacles. 
Our adaptation approach encourages the inherent regularities on the estimated maps to guide the agent to overcome the prevalent domain discrepancy in a novel environment.
Specifically, we propose an efficient learning curriculum to handle the visual and dynamics corruptions in an online manner, self-supervised with pseudo clean maps generated by style transfer networks.
Because the map-based representation provides spatial knowledge for the agent's policy, our formulation can deploy the pretrained policy networks from simulators in a new setting.
We evaluate MoDA in various practical scenarios and show that our proposed method quickly enhances the agent's performance in downstream tasks including localization, mapping, exploration, and point-goal navigation.

\end{abstract}

\section{Introduction}

The absence of ground-truth labels is a critical bottleneck for training embodied agents in complex 3D world.
A widely-used alternative is to train the agents in interactive simulators~\cite{habitat19iccv,szot2021habitat} which can load various 3D indoor scenes ~\cite{habitat19iccv,szot2021habitat,chang2017matterport3d}.
Yet, when the agent optimized for a simulator is deployed in the real world, it fails to persist its performance due to the various unseen environmental noises~\cite{kadian2020sim2real}. 
The domain gap between simulators and the real world may be diminished by modeling the environmental noises~\cite{chaplot2020learning, kadian2019we}, but it is not possible to obtain the correct noise model for the countless combinations of practical set-ups. 
Nonetheless, the visual agents collect an enormous amount of unlabelled data from 3D scenes over spatial movement.
If an agent can transfer its performance utilizing such data, the adaptation scheme can serve as a generic solution for an embodied agent to serve in diverse real environments.

Many studies in embodied agents have shown how the map-based memory aids an agent for robust visual navigation~\cite{chaplot2020object, karkus2021differentiable, wani2020multion}. 
The agent aligns its egocentric visual observations to generate a top-down map representing the environment's layout. 
The allocentric understanding helps the agent to localize itself and plan for various navigation tasks efficiently. 
Furthermore, the map provides a domain-agnostic representation, disentangling the agent's perceptual module from its planning. 
In a scenario where the pretrained agent is transferred to a new, noisy environment with various visual and dynamics corruptions, we suggest a domain adaptation method which only fine-tunes the domain-agnostic map memory, rather than the overall pipeline of the embodied agent. 

\begin{figure}[t]
\centering
\includegraphics[width=0.87\linewidth]{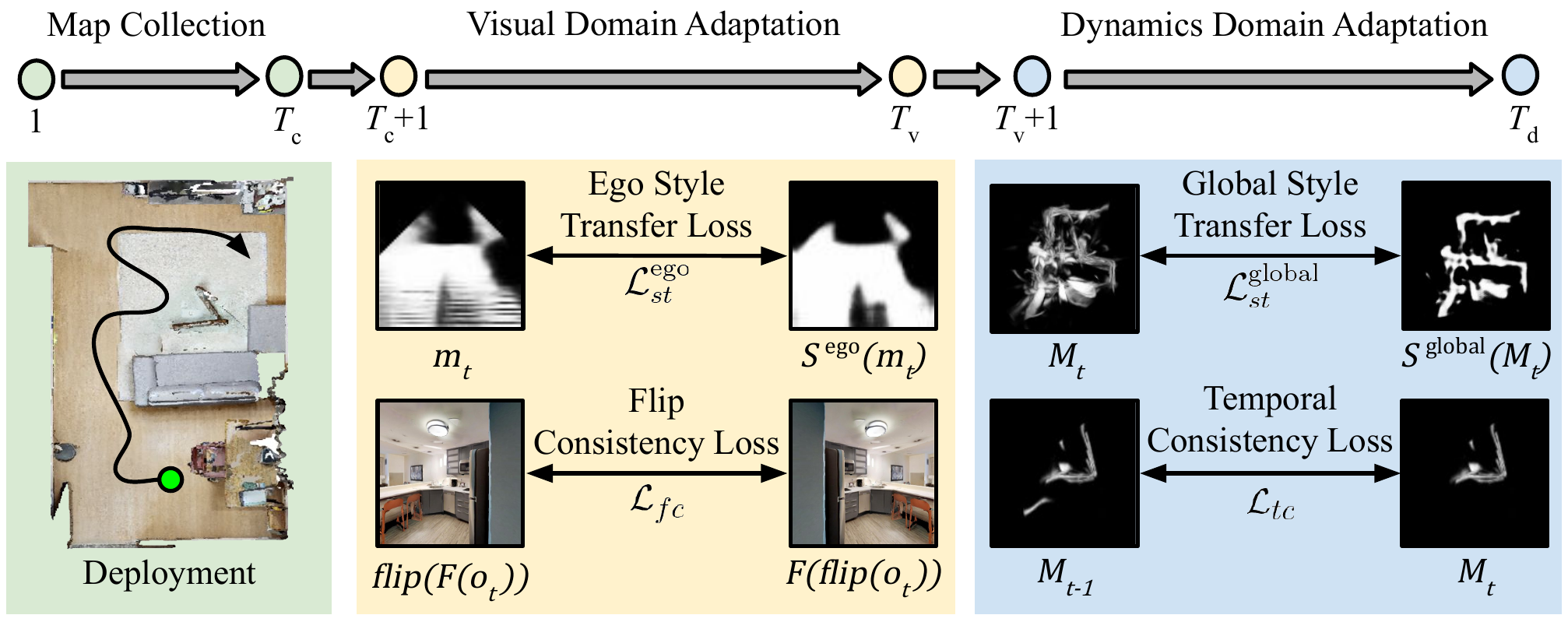}
\caption{MoDA suggests an integrated domain adaptation method for visual and dynamics corruptions, which fine-tunes an agent in an efficient learning curriculum. 
The agent collects a new map dataset to learn map style transfer networks (\textit{green}).
The agent is then trained for the new visual domain with ego style transfer loss and flip consistency loss (\textit{yellow}).
Lastly, the dynamics domain adaptation transfers the agent with global style transfer loss and temporal consistency loss (\textit{blue}).
}
\label{fig:curriculum_learning}
\end{figure}

To compensate for the absence of ground-truth in the real world, we propose a self-supervised domain adaptation method, MoDA.
The proposed method transfers the pretrained agent to the new, noisy environment by learning style transfer networks on maps. 
Our objective is to learn the structural regularities of indoor scenes from the clean maps obtained from the noiseless simulator.
We then transfer the style onto the new maps generated amidst visual and dynamics noises. 
Our self-supervision loss compares the generated maps with the style-transferred maps. 
More specifically, our method suggests a learning curriculum as shown in Fig.~\ref{fig:curriculum_learning}.
First, the agent is deployed to collect the noisy maps and learns two style transfer networks for egocentric and global maps. 
We then transfer the agent for the visual corruptions through the ego style loss, followed by compensating for the dynamics corruptions with the global style loss.  
To stabilize the training, we additionally encourage the flip consistency on RGB observations and temporal consistency over the agent's movement. 
MoDA provides an integrated self-supervised solution for both visual and dynamics corruptions and enables online adaptation in an environment where the ground-truth is unavailable. 
 
We analyze MoDA in various domain adaptation scenarios where 
the pretrained agent is transferred to the novel environments with both visual and dynamics corruptions. 
We investigate multiple types of visual corruptions along with the two main dynamics corruptions, which are the odometry sensor noise and actuation noise.
Our experiments show that the proposed adaptation method effectively enhances the embodied agents' performance in localization, mapping, and the downstream navigation tasks. 
To summarize, our main contributions are as follows: 
i) we propose a self-supervised domain adaptation method using map style transfer,
ii) we suggest an efficient curriculum to learn an adaptation integrated for visual and dynamics corruptions, and
iii) we demonstrate that the proposed approach enhances the agent's performance in localization, mapping, and the final downstream visual navigation tasks in novel, unseen environments.
\section{Related Work}

In this section, we describe existing approaches for our main task, visual navigation and simulation-to-reality adaptation (Sim2Real), along with methods for image translation which is the key technique of our work.

\paragraph{Visual Navigation}
The objective of visual navigation is to devise vision-based mapping and planning policies for solving a designated task.
Classical approaches generate a map of the environment using SLAM techniques~\cite{cadena2016past,lsd_slam,durrant2006simultaneous}, and apply planning algorithms~\cite{koenig2002d,wang2011application,Hart1968} using the generated map.
On the other hand, recent learning-based approaches often train agents end-to-end with integrated mapping and planning~\cite{chaplot2020learning, chaplot2020neural, chaplot2020object, ramakrishnan2020occupancy}.
These agents have shown competitive performance in a wide variety of tasks such as embodied QA~\cite{das2018embodied} and goal-oriented navigation~\cite{chaplot2020object, ramakrishnan2020occupancy}.

Maintaining a dedicated spatial memory unit is a key to such learning-based navigation agents.
The spatial structure of the surrounding environment is often implicitly encoded with LSTM or GRU~\cite{HochSchm97, chung2014empirical}, or using graph structures that embed keyframes as graph nodes~\cite{chaplot2020neural,chen2019behavioral,deng2020evolving,savinov2018semi}.
Nevertheless, map-based memories that depict spatial information on occupancy grid maps~\cite{chaplot2020learning, chaplot2020object, ramakrishnan2020occupancy,chaplot2021seal} efficiently aid embodied agents for tasks requiring long-range tracking and spatially-grounded planning.
We mainly retain our focus on map-based spatial memory and propose a self-supervised task formulated on grid maps for effective domain adaptation.

\paragraph{Sim2Real}
Simulators enable training an embodied agent with ground-truth poses or labels.
While recent simulators~\cite{habitat19iccv,szot2021habitat} can realistically model the world to a certain degree, there are non-idealities in agent and object dynamics.
More importantly, domain gap is inevitable when deploying an agent trained in simulation to real-world environments.
As a result, agents trained on simulators often fail to generalize in real-word settings~\cite{kadian2020sim2real}.
To alleviate Sim2Real gap, domain randomization~\cite{Chattopadhyay2021RobustNavTB,domrand} proposes to train the agent in various dynamics and visual simulations, which in turn allows the agent to observe a wide range of domains before actual deployment.
Furthermore, representation learning techniques are used to improve the generalization performance in the context of embodied agent studies~\cite{fan2021secant, hansen2021generalization, shah2021rrl, srinivas2020curl, du2021curious}. 
Alternatively, Hansen et al.~\cite{hansen2020self} proposes to adapt the policy during real-world deployment, using self-supervised objectives such as inverse dynamics and rotation prediction.
Recently, Lee et al.~\cite{Lee_2022_WACV} introduced a self-supervised domain adaptation algorithm that is formulated upon occupancy maps, which showed performance enhancement in various deployment scenarios.
However, Lee et al.~\cite{Lee_2022_WACV} requires multiple agent round trips for successful adaptation, which limits the practical usage of the algorithm.

We compare MoDA against existing approaches for adapting agents to Sim2Real deployment and demonstrate that MoDA can perform effective adaptation without mandating fixed agent trajectories such as round trips.

\paragraph{Image-to-Image Translation}
The goal of image-to-image translation is to transfer an image from a source domain into the style of target domain but to maintain its key contents.
Early approaches such as Pix2Pix~\cite{pix2pix2017} propose to use generative adversarial networks (GANs)~\cite{gan} for paired image-to-image translation.
CycleGAN~\cite{CycleGAN2017} aims to solve a more challenging problem of unpaired translation, where only a group of source and target domain images are provided for training.
To accommodate for the lack of paired data, CycleGAN~\cite{CycleGAN2017} proposes a novel cycle consistency loss that learns a forward and backward mapping simultaneously, leading to realistic image transfer.
We leverage CycleGAN for transforming occupancy grid maps to the target domain, which allows for effective map generation under new, unseen environments.
While recent advances in image-to-image translation enable high-resolution or multi-modal synthesis~\cite{wang2018pix2pixHD,zhu_multimodal,huang2018munit}, we find that CycleGAN~\cite{CycleGAN2017} is sufficient for transferring between occupancy grid maps that is not as diverse as real-world images.

\section{Method}
Given a pretrained agent from a noiseless simulator, MoDA transfers the agent to unseen, noisy environments with visual and dynamics corruptions.
We first describe the overall pipeline of visual navigation with map-based memory in Sec.~\ref{sec:visual_navigation}.
Then Sec.~\ref{sec:style_transfer} describes our map-to-map style transfer network which serves as the self-supervision signal.
Lastly, Sec.~\ref{sec:curriculum_learning} provides the learning curriculum of our online domain adaptation.
\begin{figure}[t]
\centering
\includegraphics[width=\columnwidth]{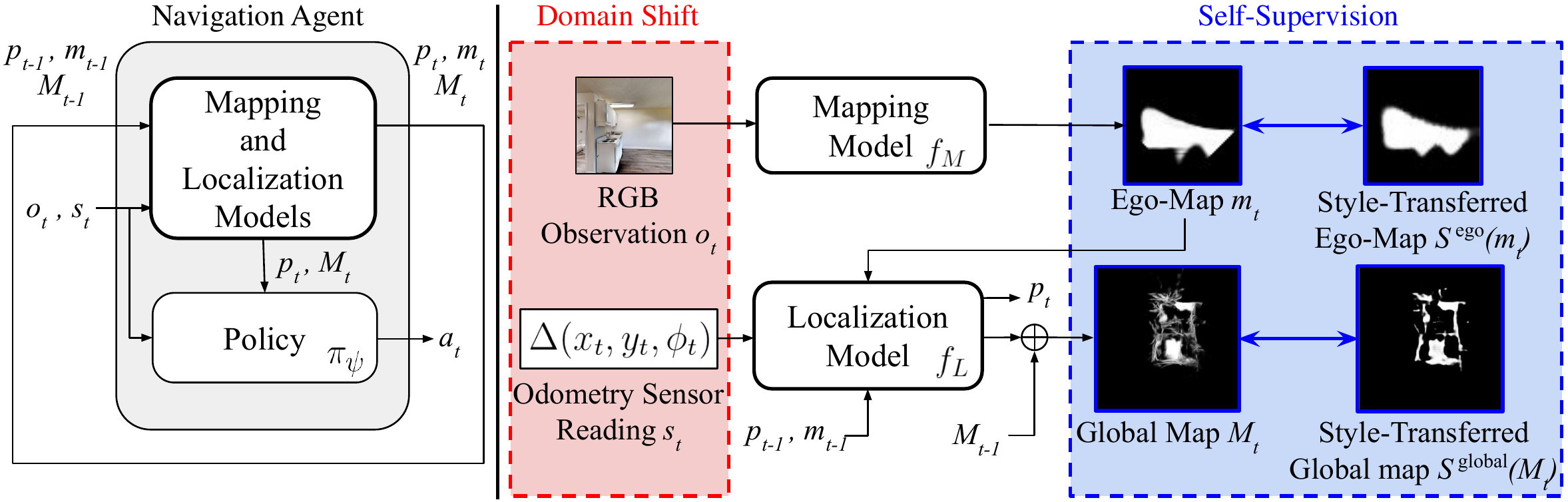}
\caption{The structure of map-based navigation agent decouples the mapping and localization models from the policy with the intermediate map memory (\textit{left}). We suggests a domain adaptation method which transfers an agent to a shifted domain with visual and dynamics corruptions (\textit{right}). Given the corrupted sensory inputs, $o_{t}$ and $s_{t}$, our method only fine-tunes the agent's mapping $f_{M}$ and localization $f_{L}$ models with the self-supervision loss, encouraging the generated ego-map $m_{t}$ and global map $M_{t}$ to be similar to the style-transferred maps.  
}
\label{fig:navigation_agent}
\end{figure}

\subsection{Visual Navigation with Spatial Map Memory}
\label{sec:visual_navigation}
MoDA builds on conventional map-based navigation agents, where the action policy is planned based on map representation as shown in Fig.~\ref{fig:navigation_agent}.
The global map is estimated from the mapping and localization models.
At each step, the RGB observation ${o}_t$ is given as an input to the mapping model $f_M$ which predicts the egocentric map $m_t$:
\begin{align}
    f_{M}(o_{t}) = m_{t}.
\end{align}
Given the odometry sensor measurement $s_t$, the localization model $f_L$ predicts the 2D pose, $p_t=(x,y,\phi)$, where
$(x,y)$ indicates the 2D coordinate and $\phi$ denotes the 1D orientation.
The localization model $f_L$ is represented as 
\begin{align}
    f_{L}(p_{t-1}, s_{t}, \{m_{t-1}, m_{t}\}) =   {p}_{t}.
\end{align}
Note that the predicted egocentric maps from the previous and current timesteps $\{m_{t-1}, m_t\}$ are given to obtain the pose prediction $\Delta p_t = p_t - p_{t-1}$. 
Then the egocentric maps from the mapping model are transformed by the estimated poses from the localization model and accumulated as a global map $M_t$:
\begin{align}
    M_{t-1} \oplus \mathcal{T}_{p_t}(m_{t})  = M_{t},
\end{align}
where $\oplus$ represents the fusion of 2D grid maps.
The representation of transformation is simplified in Fig.~\ref{fig:navigation_agent}.
Then the policy module $\pi_\psi$ plans an action $a_t=(u_x, u_y, u_\phi) \in \mathcal{A}$. 
The policy is derived from the sensory inputs $o_{t},s_{t}$, the agent's current pose ${p}_t$ and the global map $M_{t}$,
\begin{align}
    \pi_{\psi}(o_{t},s_{t},{p}_{t},M_{t}) = a_{t}.
\end{align}
Nonetheless, many studies have shown that the policy is most dependent on the map-based memory $M_t$, which is useful for long-range or complex navigation, rather than the sensory inputs providing partial observability~\cite{wani2020multion,bhatti2016playing,gupta2017cognitive}. 


The agent performs various tasks including mapping and localization using the policy module trained in an ideal environment with ground-truth poses and maps.
However, in realistic environments, the overall architecture is disturbed by two main types of corruption: visual and dynamics corruptions.
The visual corruptions on RGB observations make it difficult to predict egocentric maps while the dynamics corruptions on odometry readings and actuation degrade the pose estimation. 
As the modular pipeline of map-based agents decouples the policy from mapping and localization models with the intermediate spatial memory, MoDA only fine-tunes the perceptual models, namely localization and mapping, to learn the perturbations in measurements and generate a domain-agnostic map for the subsequent policy.

\paragraph{Visual Corruptions}
Visual corruptions affect the RGB observation, which is the input of the mapping model shown in Fig.~\ref{fig:navigation_agent}.
As a result, the egocentric perception of the embodied agent suffers from the domain discrepancy.
While our adaptation does not assume any particular form of visual corruption, we test our adaptation in scenarios that properly represent the wide varieties of possible visual variations in the real world~\cite{Chattopadhyay2021RobustNavTB}.
The tested scenarios are described in Sec.~\ref{sec:experiments}.

\paragraph{Dynamics Corruptions}
Dynamics corruptions degrade the accuracy of the agent's pose estimated by the localization model, as shown in Fig.~\ref{fig:navigation_agent}.
The two main sources of dynamics corruptions are the actuation and odometry sensor noises.
The actuation noise interrupts an agent from reaching the target location provided by the control commands. 
After an action, the agent's ground-truth movement $\Delta p_t=\Delta (x_t, y_t, \phi_t)$ is defined as 
\begin{align}
    \Delta (x_{t},y_{t},\phi_{t}) = (u_x,u_y,u_\phi) + \epsilon_\text{act},
    \label{eq:pose_actuation_noise}
\end{align}
where $(u_x,u_y,u_\phi)$ and $\epsilon_\text{act}$ indicate the intended action control and the actuation noise, respectively.  
Additionally, the odometry sensor noise $\epsilon_\text{sen}$ disturbs the agent from accurately perceiving its own movement. 
The final pose reading with the odometry sensor noise $\epsilon_\text{sen}$ is erroneously measured as 
\begin{align}
   s_t= \Delta (x_{t},y_{t},\phi_{t}) + \epsilon_\text{sen}. 
   \label{eq:pose_odometry_noise}
\end{align} 
The actuation and odometry sensor noises are expected in all realistic settings and many studies present realistic models for both types of dynamics corruptions~\cite{gonccalves2008sensor,khosla1989categorization, habitat19iccv, Chattopadhyay2021RobustNavTB}.

\subsection{Unpaired Map-to-map Translation Network}
\label{sec:style_transfer}
The main objective of our self-supervised domain adaptation is to translate maps observed from the new, noisy domain such that it recovers the noiseless structure in the absence of paired data.
The neural network learns to capture the structural regularities of indoor scenes from the collection of ground-truth maps $D_\text{gt}$ and translate the learned characteristics into the collection of noisy maps $D_\text{noisy}$. 
The set of ground-truth maps $D_\text{gt}$ is collected from a noiseless simulator.
When the pretrained agent is deployed in a new environment, it collects another set of map data $D_\text{noisy}$ which is generated amidst visual and dynamics corruptions.
We adopt the unpaired image-to-image translation network suggested in~\cite{CycleGAN2017} on our maps.
Since the formulation does not require one-to-one correspondences between the two sets, our domain adaptation is completely self-supervised and successfully reasons about the stylistic difference between the two collections.

\begin{figure}[t]
\centering
\includegraphics[width=.98\columnwidth]{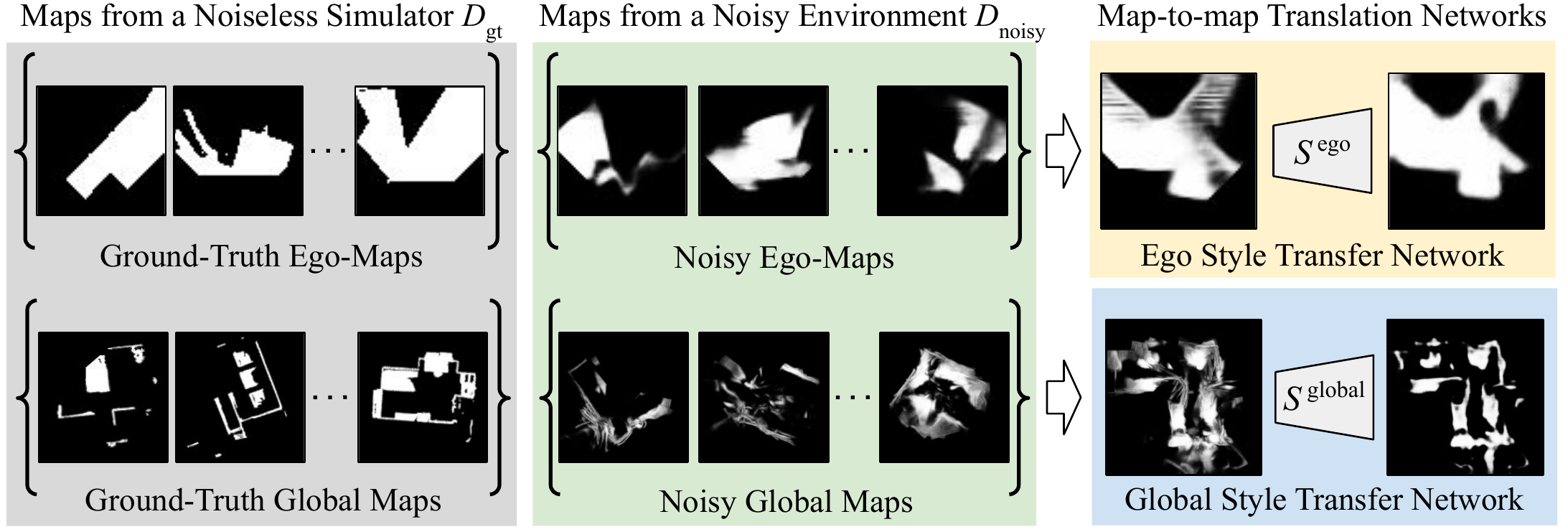}
\caption{Given the set of ground-truth maps $D_\text{gt}$(\textit{grey}) obtained from the noiseless simulator, the set of noisy maps $D_\text{noisy}$(\textit{green}) is collected by the pretrained agent deployed in a novel environment amidst visual and dynamics corruptions. 
We then learn two map-to-map translation networks for the egocentric map $S^\text{ego}$ (\textit{yellow}) and global map $S^\text{global}$ (\textit{blue}) to translate the maps in $D_\text{noisy}$ into the style of the maps in $D_\text{gt}$}
\label{fig:style-transfer}
\end{figure}

More specifically, we learn two map-to-map translation networks for the egocentric map $S^\text{ego}$ and the global map $S^\text{global}$ as shown in Fig.~\ref{fig:style-transfer}. 
The egocentric maps observe magnified views of the environment, and the style transfer network $S^\text{ego}$ can help the agent to learn the detailed visual structure of the indoor scene to train the mapping model $f_M$ against visual corruptions. 
The global style transfer network $S^\text{global}$, on the other hand, enforces a globally coherent structure over long-horizon navigation.
The self-supervised loss on global maps restricts the localization model $f_L$ from making erroneous predictions due to dynamics corruptions. 

The success of the style transfer loss is tightly coupled with the clear structural regularities within the desired set of map data, which are represented as simple gray-scale images.
We use different representations that contain more information in respective scenarios;
the network learns over the explored area for egocentric maps, whereas the obstacle maps are used for global maps.
While the two different representations are easily converted from one to the other given the pose of the agent, the obstacle maps are much more sparse than the explored area.
Because the ego-maps cover smaller region, the obstacle maps cannot produce meaningful prior,  consisting only about 6\% of nonzero values on average within the images.
On the other hand, the large overlapping regions in explored areas can be challenging in noisy global maps, whereas obstacle maps exhibit clearly distinguished structure.

\subsection{Curriculum Learning for Domain Adaptation} 
\label{sec:curriculum_learning}

We design a sequential curriculum using the hierarchical structure of map-based models as shown in Fig.~\ref{fig:curriculum_learning}.
Once the agent collects the map data during its initial deployment up to time $T_{c}$, it then learns the two style transfer networks for the new environment.
In the next step, our self-supervised adaptation method fine-tunes the perceptual module to robustly handle unknown corruptions.
The style transfer loss on the egocentric maps provides signals to adapt the mapping model $f_M$ against visual corruptions.
Additionally, we enforce flip consistency on the RGB observations. 
Next, the global style transfer loss fine-tunes the localization model $f_L$ up to time $T_d$ against dynamics corruptions, in addition to the temporal consistency in the global map.
The transferred agent then stably performs various navigation tasks in the new, noisy environment.

\paragraph{Visual Domain Adaptation}
According to the learning curriculum, we first adapt the mapping model for unknown visual changes in the new environments. 
The style transfer network for the egocentric map $S^\text{ego}$ converts the predicted egocentric map $m_t$ into noiseless style.
The ego style transfer loss minimizes the discrepancy between the two maps:
\begin{align}
   \mathcal{L}^\text{ego}_{st} & = \sum_{t=T_{c}+1}^{T_{v}} \| m_{t} - S^\text{ego}(m_{t}) \|_{2},
\end{align}
with $T_{v}$ indicating the ending time of visual domain adaptation.
In addition, we fine-tune the feature extractor $F$ of the mapping model $f_M$ along with a consistency loss to stabilize the training. 
The flip consistency loss $\mathcal{L}_{fc}$ assumes that the feature extractor should make consistent predictions over the flipped observations~\cite{li2021self, Araslanov_2021_CVPR}. 
Specifically, when a horizontally flipped RGB observation, $\text{flip}(o_{t})$, and a non-flipped original observation, $o_{t}$, are given as inputs, the estimated egocentric maps should be equal but flipped.
We, therefore, define our flip consistency loss as 
\begin{align}
   \mathcal{L}_{fc} & = \sum_{t=T_{c}+1}^{T_{v}} \| flip(F(o_{t})) - F(flip(o_{t})) \|_{2},
\end{align}
Together the visual domain adaptation transfers the mapping model of the pretrained agent with the visual domain loss
\begin{align}
    \mathcal{L}_{V} &= \lambda^\text{ego}_{st} \mathcal{L}^\text{ego}_{st} + \lambda_{fc} \mathcal{L}_{fc}.
\end{align}
The values for hyper-parameters are provided in the supplementary material.

\paragraph{Dynamics Domain Adaptation}
In the next stage, we adapt the localization model $f_L$ to the dynamics corruptions that are present in the new environment.
The agent generates a global map over its trajectory and encodes the predicted pose information onto the map. 
Thus, by learning the structural priors from the style transfer network, the agent can inversely learn to estimate more accurate pose. 
Given the estimated and style-transferred global maps, the global style transfer loss, $\mathcal{L}^\text{global}_{st}$ is formulated as
\begin{align}
  \mathcal{L}^\text{global}_{st} & = \sum_{t=T_{v}+1}^{T_{d}} \| M_{t} - S^\text{global}(M_{t}) \|_{2}.  
\end{align}
where $T_{d}$ denotes the ending time of dynamics domain adaptation.
Moreover, we encourage the agent to generate consistent global maps over time. 
As the pose error accumulates over time, the global map generated in the earlier step encodes more accurate pose information over the global map generated in the later step. 
The temporal consistency loss $\mathcal{L}_{tc}$ compares the generated global map to the map from the previous time-step, and it is defined as
\begin{align}
   \mathcal{L}_{tc} & = \sum_{t=T_{v}+1}^{T_{d}} \| M_{t} - M_{t-1} \|_{2}.
\end{align}
Therefore, the full objective of dynamics domain adaptation becomes 
\begin{align}
    \mathcal{L_{D}} &= \lambda^\text{global}_{st} \mathcal{L}^\text{global}_{st} + \lambda_{tc} \mathcal{L}_{tc},
\end{align}
completing the final stage of suggested learning curriculum. 

\section{Experiments}
\label{sec:experiments}

\begin{figure}[t]
\centering
\includegraphics[width=0.7\linewidth]{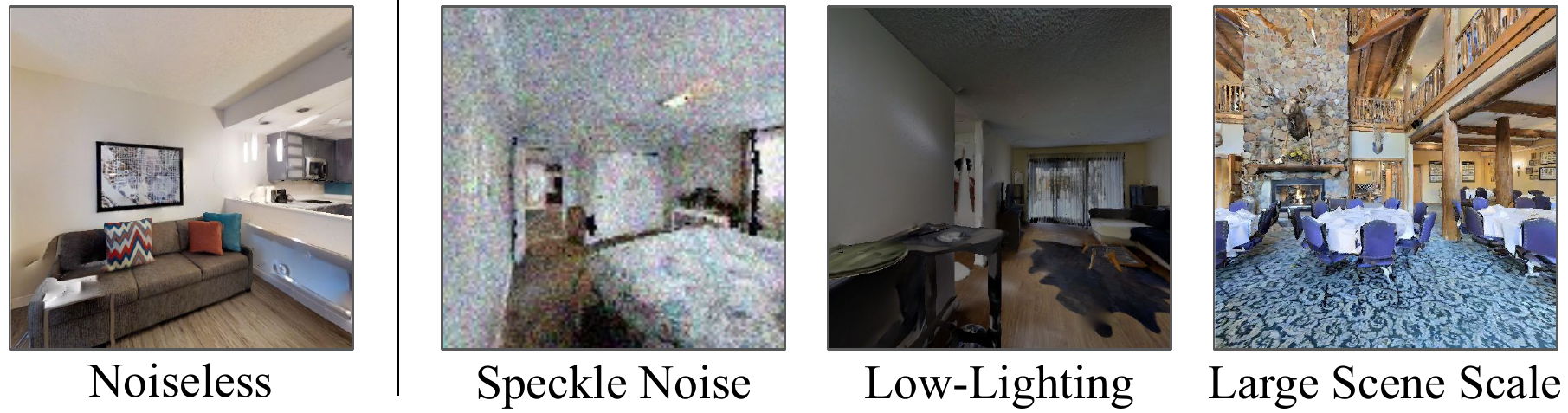}
\caption{Visualization of RGB observations with visual corruptions: speckle noise, low-lighting and large scene scale}
\label{fig:visualcorruptions}
\end{figure}

We show the validity of MoDA using the navigation agent from Active Neural SLAM~\cite{chaplot2020learning} which is widely adapted for other navigation models~\cite{chaplot2020object,ramakrishnan2020occupancy}. 
Nonetheless, MoDA is applicable to various navigation agents with map-like memory. 
We use the Habitat simulator~\cite{habitat19iccv}.
The pretrained agent is trained on the standard train split~\cite{habitat19iccv} of Gibson dataset~\cite{xia2018gibson} with ground-truth supervision.
We split the unseen scenes of Gibson and Matterport3D~\cite{chang2017matterport3d} for adaptation and evaluation. 
The scenes for each split are listed in the supplementary material. 
MoDA is implemented using Pytorch~\cite{paszke2019pytorch} and accelerated with an RTX 2080 GPU.

\paragraph{Visual and Dynamics Corruptions}
We evaluate the proposed method in three environments where visual and dynamics corruptions are present. 
Each environment is distinguished by the three visual variations: speckle noise, low-lighting, and scene scale change.
Specifically, our experiments transfer the pretrained agent in three types of variation visualized in Fig.~\ref{fig:visualcorruptions}.
First, we apply image quality degradation, which may be caused by the physical condition of the mounted camera.
We generate low-quality RGB observations with additive speckle noises.
The second type of perturbation is the low-lighting condition to reflect the common light variations in the real world.
We show if our agent can be transferred to low-lighting scenes by adjusting the contrast and brightness of the input RGB image. 
Lastly, we evaluate if our pretrained agent from Gibson scenes can generalize to different scene scale by transferring the agent to the scenes in Matterport3D. 
While the Gibson scenes consist of scans collected from offices, the scenes in Matterport3D generally consist of large-scale homes. 
As the dynamics corruptions are seen in all realistic environments, we add the odometry sensor and actuation noise models to all three scenarios. 
In our experiment, we use the noise parameters generated from the actual physical deployment of LoCobot~\cite{locobot} in previous work~\cite{chaplot2020learning,kadian2020sim2real} and draw from a Gaussian Mixture Model at each step. 

\paragraph{Baselines}
We extensively compare our agent, referred as ``\textbf{MoDA}" to various baselines.
``No adaptation (\textbf{NA})" reflects the performance degradation of the un-adapted, pretrained model due to the domain gap. 
``Domain Randomization (\textbf{DR})" adapts the pretrained model with ground-truth supervision in a randomized domain with various combinations of visual and dynamics corruptions.
``Policy Adaptation during Deployment (\textbf{PAD})" proposed by Hansen et al.~\cite{hansen2020self} performs visual domain adaptation using an auxiliary task, namely rotation prediction.
As the original method mainly targets visual adaptation, we further extend PAD for dynamics corruption by additionally training with our dynamics adaptation method.
Lastly, ``Global Map Consistency (\textbf{GMC})" from Lee et al.~\cite{Lee_2022_WACV} imposes global map consistency loss on round trip trajectories to adapt to dynamics corruptions.
Further details of baseline implementation are explained in the supplementary material.

\paragraph{Tasks and Evaluation Metric}
We report the transferred agent's localization and mapping performance in the new, noisy environment.  
For fair comparison, we evaluate each adaptation method on an identical set of trajectories obtained from the un-adapted agent in each environment.
Following~\cite{campbell2018globally,Kim_2021_ICCV}, localization is evaluated with the median translation ($x,y$) and rotation ($\phi$) error.
Mapping performance is evaluated with the mean squared error (MSE) of the generated occupancy grid maps compared to the ground-truth.

We also demonstrate our adapted agent's performance in downstream navigation tasks. 
Following Chaplot et al.~\cite{chaplot2020learning}, we report exploration performance using the explored area and explored area ratio after letting the agent to explore for a fixed number of steps. 
In addition, we report the collision ratio, which is the percentage of collisions from the agent's total steps.
We include the collision ratio to distinguish simple random policy, which often results in undesirable collisions coupled with sliding along the walls, and eventually explore large areas.
We further evaluate our agent on point-goal navigation(PointNav) as suggested in~\cite{anderson2018evaluation}. 
Here, we report the success rate and Success weighted by Path Length(SPL)~\cite{anderson2018evaluation}.

\paragraph{}
We investigate MoDA in two settings, generalization and specialization,  following ~\cite{chaplot2021seal}.
In our main experiment, \textit{generalization} adapts the agent in a set of unseen scenes with unknown noises (Sec.~\ref{sec:exp_generalization}). 
The trained agent is then evaluated in a different set of novel scenes but with the same visual and dynamics corruptions. 
For \textit{specialization}, the agent is fine-tuned and evaluated in the same set of unseen scenes, but it starts from a different initial pose for evaluation (Sec.~\ref{sec:exp_specialization}).

\begin{table}[t]
\caption{Generalization performance in the three new environments with speckle noise, low-lighting, and scene scale change. All three scenes contain dynamic corruptions not present in the original training setup of the pretrained agent}
\label{table:generalization}
\centering
\setlength{\tabcolsep}{5.5pt}
\resizebox{0.8\columnwidth}{0.4cm}{
\subfloat[Gibson Scenes Containing Speckle Noise and Dynamics Corruptions]{%
\begin{tabular}{l|c|c|c|c|c|c|c|c|c}
\toprule
\multirow{2}{*}{} & \multicolumn{2}{c|}{Pose} &\multicolumn{2}{|c|}{Map(MSE)}  &  \multicolumn{3}{|c|}{Exploration}  & \multicolumn{2}{|c}{PointNav} \\
\midrule
      &    $x,y$(m) & $\theta$($^\circ$) &  ego   & global &  area & ratio & collision & success & SPL  \\ \midrule
        NA & 0.16 & 15.02 & 1.11 & 0.32 & 28.45 & 0.82 & 0.40 & 0.12 & 0.10 \\ 
        DR & 0.13 & 8.74  & 1.14 & 0.27 & 29.35 & \textbf{0.88} & 0.43 & 0.20 & 0.13 \\ 
        PAD & \textbf{0.04}& \textbf{1.08}  & 1.35 & 0.32 & 22.83 & 0.66 & 0.51 & 0.08 & 0.07 \\ 
        GMC & 0.06 & 5.10 & 1.11 & 0.28 & \textbf{29.73} & 0.85 & \textbf{0.35} & 0.36 & 0.29 \\ 
        MoDA & \textbf{0.04} & 2.61  & \textbf{1.08} & \textbf{0.25} & 28.63 & 0.82 & 0.36 & \textbf{0.56} & \textbf{0.47} \\ 
\bottomrule
\end{tabular}}}%

\setlength{\tabcolsep}{5.5pt}
\resizebox{0.8\columnwidth}{0.4cm}{
\subfloat[Gibson Scenes under Low-Lighting and Dynamics Corruptions]{%
\begin{tabular}{l|c|c|c|c|c|c|c|c|c}
\toprule
\multirow{2}{*}{} & \multicolumn{2}{c|}{Pose} &\multicolumn{2}{|c|}{Map(MSE)}  &  \multicolumn{3}{|c|}{Exploration}  & \multicolumn{2}{|c}{PointNav} \\
\midrule
      &    $x,y$(m) & $\theta$($^\circ$) &  ego   & global &  area & ratio & collision & success & SPL  \\ \midrule
        NA &  0.18 & 15.78 & 0.90 & 0.32 & 30.31 & 0.87 & 0.34 & 0.22 & 0.17 \\ 
        DR & 0.14 & 7.60 & 0.96 & 0.26 & 29.95 & 0.88 & 0.41 & 0.20 & 0.14 \\ 
        PAD & \textbf{0.05} & \textbf{2.17} & 1.02 & 0.27 & 26.88 & 0.78 & 0.37 & 0.22 & 0.18 \\ 
        GMC &  0.06 &5.39 & 0.90 & 0.28 & \textbf{32.45} & \textbf{0.91} & 0.32 & 0.46 & 0.37 \\ 
        MoDA & \textbf{0.05} & 2.87 & \textbf{0.89} & \textbf{0.25} & 31.56 & \textbf{0.91} & \textbf{0.26} & \textbf{0.56} & \textbf{0.45} \\ 
\bottomrule
\end{tabular}}}%

\setlength{\tabcolsep}{5.5pt}
\resizebox{0.8\columnwidth}{0.4cm}{
\subfloat[MatterPort3D Scenes with Large Scene Scale and Dynamics Corruptions]{%
\begin{tabular}{l|c|c|c|c|c|c|c|c|c}
\toprule
\multirow{2}{*}{} & \multicolumn{2}{c|}{Pose} &\multicolumn{2}{|c|}{Map(MSE)}  &  \multicolumn{3}{|c|}{Exploration}  & \multicolumn{2}{|c}{PointNav} \\
\midrule
      &    $x,y$(m) & $\theta$($^\circ$) &  ego   & global &  area & ratio & collision & success & SPL  \\ \midrule
        NA & 0.17 &14.42 &  1.07 & 0.39 & 52.16 & 0.45 & 0.44 & 0.04 & 0.02 \\ 
        DR &  0.14 & 9.17 & 1.10 & 0.35 & 59.22 & 0.50 & 0.43 & 0.08 & 0.06 \\ 
        PAD & \textbf{0.05} & 3.12 & 1.26 & 0.41 & 41.66 & 0.35 & 0.49 & 0.02 & 0.02 \\ 
        GMC  & 0.10& 6.05 & 1.07 & 0.40 & 54.73 & 0.47 & 0.36 & 0.10 & 0.08 \\ 
        MoDA & \textbf{0.05} & \textbf{2.34} & \textbf{1.02} & \textbf{0.31} & \textbf{63.68} & \textbf{0.54} & \textbf{0.28} & \textbf{0.22} & \textbf{0.18} \\ 
\bottomrule
\end{tabular}}}%
\end{table}

\begin{figure}[t]
\centering
\includegraphics[width=\columnwidth]{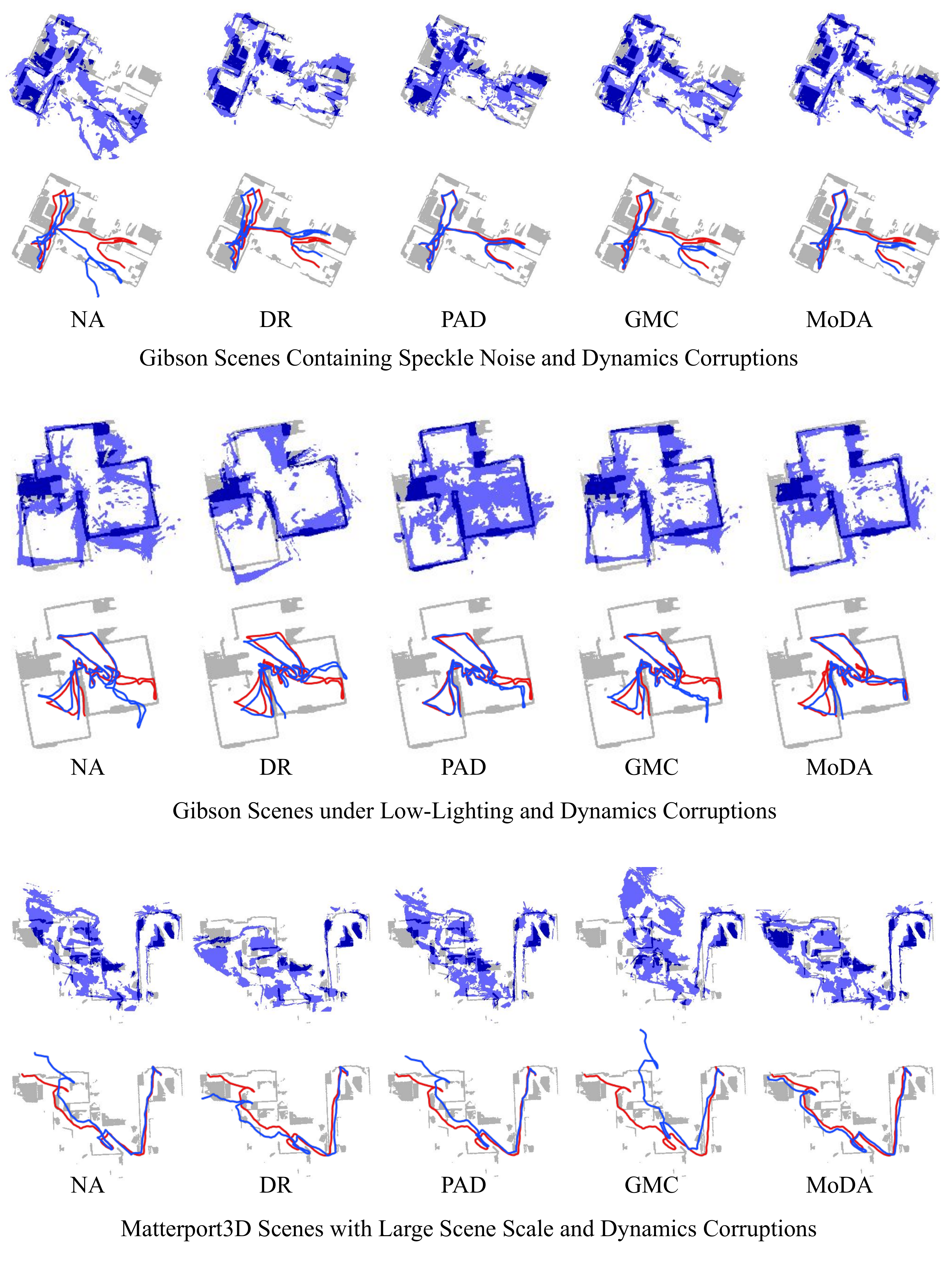}
\caption{
Qualitative result of mapping (\textit{top}) and localization (\textit{bottom}) obtained from agents observing the identical sequence of RGB observations and odometry sensor readings. The reconstructed maps (\textit{blue}) are aligned on the ground-truth maps (\textit{grey}), and the estimated pose trajectories (\textit{blue line}) are compared to the ground-truth trajectories (\textit{red line}) 
}
\label{fig:qual_result}
\end{figure}
\begin{table}[t]
\caption{Specialization result in the three new environments with speckle noise, low-lighting, and scene scale change. All scenes contain dynamics corruptions}
\label{table:specialization}
\resizebox{\columnwidth}{!}{
\begin{tabular}{llcc|cc|cclcc|cc|cclcc|cc|cc}
\toprule
     &  & \multicolumn{6}{c}{Gibson Speckle Noise}      &  & \multicolumn{6}{c}{Gibson Low-Lighting}           &  & \multicolumn{6}{c}{Matterport3D Large Scene Scale}                      \\ \cline{3-8} \cline{10-15} \cline{17-22} 
 &
   &
  \multicolumn{2}{c|}{Pose} &
  \multicolumn{2}{c|}{Map(MSE)} &
  \multicolumn{2}{c}{PointNav} &
   &
  \multicolumn{2}{c|}{Pose} &
  \multicolumn{2}{c|}{Map(MSE)} &
  \multicolumn{2}{c}{PointNav} &
   &
  \multicolumn{2}{c|}{Pose} &
  \multicolumn{2}{c|}{Map(MSE)} &
  \multicolumn{2}{c}{PointNav} \\ \cline{3-8} \cline{10-15} \cline{17-22} 
     &  &     $x,y$(m) & $\theta$($^\circ$)  & ego  & global & success & SPL  &  &    $x,y$(m) & $\theta$($^\circ$)  & ego  & global & success & SPL  &  &     $x,y$(m) & $\theta$($^\circ$)  & ego  & global & success & SPL  \\ 
\midrule
NA&  &  0.17 &15.25 & 1.10 & 0.30   & 0.18 & 0.16 &   & 0.17& 14.44 & 0.87 & 0.29   & 0.18 & 0.17 &   & 0.16 & 15.04& 1.09 & 0.38   & 0.04    & 0.03 \\
DR&   & 0.13 & 9.17 & 1.14 & 0.28   & 0.20 & 0.16 &   & 0.14& 9.11  & 0.92 & 0.27   & 0.22 & 0.17 &    & 0.13& 9.54 & 1.11 & 0.34   & 0.02    & 0.01 \\
PAD  &  & \textbf{0.03} &\textbf{1.09}  &  1.33 & 0.31   & 0.06 & 0.05 &  &  \textbf{0.04} &\textbf{2.88}  & 1.01 & 0.26   & 0.32 & 0.28 &  & 0.05 & 3.21  & 1.25 & 0.40   & 0.06    & 0.05 \\
GMC  &    & 0.06 & 5.66 & 1.09 & 0.28   & 0.44 & 0.38 &    & 0.07 & 7.82 & 0.87 & 0.28   & 0.42 & 0.37 &    & 0.09 & 6.51 & 1.09 & 0.38   & 0.12    & 0.08 \\
MoDA&   & 0.04 & 2.54 & \textbf{1.08} & \textbf{0.25}   & \textbf{0.56} & \textbf{0.47} &  & 0.06 & 3.24  & \textbf{0.85} & \textbf{0.25}   & \textbf{0.54} & \textbf{0.47} &    & \textbf{0.04} & \textbf{2.21} & \textbf{1.03} & \textbf{0.31}   & \textbf{0.22}    & \textbf{0.17} \\
\bottomrule
\end{tabular}
}
\end{table} 

\subsection{Task Adaptation to Noisy Environments: Generalization}
\label{sec:exp_generalization}
In generalization, we test whether the adaptation method properly transfers the agent to the existing visual and dynamics corruptions, and avoids over-fitting the agents to the particular scenes.
We compare our agent transferred to the three new environments with MoDA, as shown in Table~\ref{table:generalization}.
MoDA shows significant performance improvements in all three environments.
By effectively fine-tuning pretrained agents using style-transfer in the map domain, MoDA improves localization and mapping, further enhancing the downstream task performance in pointNav and exploration.
We additionally report the collision ratio to investigate our agent's stability in exploration.
Compared to the baselines, agent trained by MoDA distinctively reports the low number of collision ratio during its exploration steps.
While NA and DR agents also show competitive exploration performance, they also exhibit high collision ratios, which indicate instability.
Further, while GMC shows competitive performance against MoDA, it mandates the agent to navigate in round trip trajectories for adaptation.
MoDA performs successful adaptation without such constraints, thus more practical than GMC.
As a result, our agent's performance across all evaluation metrics confirms the effectiveness of MoDA which successfully adapts to a new, shifted domain with visual and dynamics corruptions. 

In Fig.~\ref{fig:qual_result}, we show the visualization of the estimated pose trajectory and reconstructed global maps generated by the agents observing the same sequence of RGB observations and odometry sensor readings.
Our model better aligns with the ground-truth compared to the baselines. 
MoDA compensates for both visual and dynamics corruptions online without using ground-truth data.

\subsection{Task Adaptation to Noisy Environments: Specialization}
\label{sec:exp_specialization}
The specialization setting reflects the practical scenario where the agent is continuously deployed in the same scenes. 
In Table~\ref{table:specialization}, we compare our agent's performance to baselines in localization, mapping, and PointNav. 
As in generalization, we evaluate the models in three different environments where the dynamics corruptions and one of each visual corruptions are present.

MoDA successfully adapts in the specialization scenario by showing coherent performance in localization, mapping, and PointNav.
In localization, our model outperforms the baselines except for PAD. 
Although PAD exhibits the lowest pose estimation error in localization metric when evaluated on logged trajectories, it fails to outperform our model in mapping or PointNav. 
GMC shows the performance enhancement in all metrics over the other baselines, yet underperforms compared to our model. 
While adapted and deployed in the same scenes, our agent stably adapts to the visual corruption as shown from the ego-map prediction result in all three environments. 
We also observe that our agent estimates distinctively accurate poses, leading to generating an accurate global map. 
The performance improvement in mapping and localization, which is tightly coupled to the corruptions added to the observation, aids our agent to generate a more accurate intermediate spatial map. 
Then, the enhanced domain-agnostic representation further leads our model to outperform the baselines for the PointNav task. 
Therefore we conclude that MoDA provides a powerful, integrated adaption method for closing the domain gap from visual and dynamics corruptions present in a realistic environment.


\begin{table}[t]
\caption{Ablation study on various loss functions employed in MoDA}
\label{table:ablation}
\centering

\resizebox{0.73\columnwidth}{!}{
\setlength{\tabcolsep}{4pt}
\begin{tabular}{l|c|c|c|c|c|c}
    \toprule
\multirow{2}{*}{} & \multicolumn{2}{c|}{Pose} &\multicolumn{2}{|c|}{Map(MSE)}  & \multicolumn{2}{|c}{PointNav} \\
\hline
      &   $x,y$(m) & $\theta$($^\circ$)  &  ego   & global & success & SPL  \\ \hline
        NA  & 0.16 & 15.02 & 1.11 & 0.32 & 0.12 & 0.10 \\ 
        NA + $\mathcal{L}_{fc}$ & 0.16  & 15.01 & 1.12 & 0.32 & 0.20 & 0.16 \\ 
        NA + $\mathcal{L}_{fc}$ + $\mathcal{L}^\text{ego}_{st}$ & 0.16 & 14.98 & \textbf{1.08} & 0.31 & 0.20 & 0.17 \\ 
        NA + $\mathcal{L}_{fc}$ +  $\mathcal{L}^\text{ego}_{st}$ +  $\mathcal{L}_{tc}$  & \textbf{0.04} & 3.16 & \textbf{1.08} & 0.25 & 0.42 & 0.34 \\ 
        NA + $\mathcal{L}_{fc}$ +  $\mathcal{L}^\text{ego}_{st}$ + $\mathcal{L}_{tc}$ + $\mathcal{L}^\text{global}_{st}$  & \textbf{0.04} & \textbf{2.61}& \textbf{1.08} & \textbf{0.25} & \textbf{0.56} & \textbf{0.47} \\ 
        \bottomrule
\end{tabular}
    }
\end{table}

\subsection{Ablation study}
\label{sec:ablation}
In this section, we verify the effectiveness of each loss function in visual and dynamics domain adaptation. 
In Table~\ref{table:ablation}, we report the adaptation result of the agent transferred to the environment with speckle noise visual corruption and dynamics corruptions.
The overall experiment setup is the same as generalization. 
Beginning from the pretrained agent, referred as ``NA", we gradually add the four losses mentioned in Sec.~\ref{sec:curriculum_learning}. 
We first train the pretrained agent only with the flip consistency loss $\mathcal{L}_{fc}$, which improves the agent's performance in localization and PointNav, but not in mapping. 
However, the model trained with both flip consistency loss $\mathcal{L}_{fc}$ and ego style transfer loss $\mathcal{L}^\text{ego}_{st}$ results in the performance enhancement in all evaluation metrics. 
This ablated model with $\mathcal{L}_{fc}$ and $\mathcal{L}^\text{ego}_{st}$ indicates that style transfer loss is more effective in adapting the agent during the visual domain adaptation stage. 
The adaptation for visual perturbations, which targets at predicting more accurate egocentric maps, also leads to the improvement in the subsequent evaluation tasks, localization, global map prediction and PointNav performance. 
The visually adapted agent is then trained with the temporal consistency loss $\mathcal{L}_{tc}$. 
The addition of $\mathcal{L}_{tc}$ effectively transfers the agent for the dynamics corruptions. 
Nonetheless, our full model, jointly trained with the global style transfer loss $\mathcal{L}^\text{global}_{st}$, exhibits the best performance in all metrics compared to all versions of ablated model.



\section{Conclusion}
In conclusion, we propose MoDA, a self-supervised domain adaptation method which provides an integrated solution to adapt the pretrained embodied agents to visual and dynamics corruptions. 
By transferring the noisy maps into clean-style maps, the agent can successfully adapt to the new environment with additional assistance with consistency loss.
Our evaluation in generalization and specialization proves that MoDA is a powerful and practical domain adaptation method, showing its applicability in the noisy real world in the absence of ground-truth.

\paragraph{\normalfont \textbf{Acknowledgements:}} This work was partly supported by the National Research Foundation of Korea (NRF) grant funded by the Korea government(MSIT) (No. 2020R1C1C1008195), Creative-Pioneering Researchers Program through Seoul National University, and Institute of Information \& communications Technology Planning \& Evaluation (IITP) grant funded by the Korea government(MSIT) (No.2021-0-02068, Artificial Intelligence Innovation Hub).

\clearpage

%
%
\bibliographystyle{splncs04}
\bibliography{egbib}

\end{document}